\documentclass{article} 
\usepackage{iclr2024_conference,times}

\usepackage{amsmath,amsfonts,bm}

\def\eqref#1{equation~\ref{#1}}

\def\1{\bm{1}}

\DeclareMathAlphabet{\mathsfit}{\encodingdefault}{\sfdefault}{m}{sl}
\SetMathAlphabet{\mathsfit}{bold}{\encodingdefault}{\sfdefault}{bx}{n}

\DeclareMathOperator*{\argmax}{arg\,max}

\newtheorem{proposition}{Proposition}

\usepackage{hyperref}
\usepackage{url}
\usepackage{ieeetrantools}
\usepackage{enumitem}
\usepackage{graphicx}
\usepackage{subfigure}
\usepackage{subcaption}
\usepackage{cleveref}


\title{FairSISA: Ensemble Post-Processing to Improve Fairness of Unlearning in LLMs}

\author{Swanand Ravindra Kadhe, Anisa Halimi, Ambrish Rawat, Nathalie Baracaldo\\
IBM Research\\
\texttt{\{swanand.kadhe, anisa.halimi\}@ibm.com},\\ \texttt{ambrish.rawat@ie.ibm.com, baracald@us.ibm.com}
}

\iclrfinalcopy 
\begin{document}

\maketitle

\begin{abstract}
Training large language models (LLMs) is a costly endeavour in terms of time and computational resources. The large amount of training data used during the unsupervised pre-training phase makes it difficult to verify all data and, unfortunately, undesirable data may be ingested during training. Re-training from scratch is impractical and has led to the creation of the \textit{unlearning} discipline where models are modified to ``unlearn'' undesirable information without retraining. However, any modification can alter the behaviour of LLMs, especially on key dimensions such as \textit{fairness}. This is the first work that examines this interplay between unlearning and fairness for LLMs. In particular, we focus on a popular unlearning framework known as SISA [Bourtoule et al., 2021], which creates an ensemble of models trained on disjoint shards. We evaluate the performance-fairness trade-off for SISA, and empirically demsontrate that SISA can indeed reduce fairness in LLMs. To remedy this, we propose post-processing bias mitigation techniques for ensemble models produced by SISA. We adapt the post-processing fairness improvement technique from [Hardt et al., 2016] to design three methods that can handle model ensembles, and prove that one of the methods is an optimal fair predictor for ensemble of models. Through experimental results, we demonstrate the efficacy of our post-processing framework called \textit{FairSISA}.
\end{abstract}

\section{Introduction}\label{sec:introduction}
Modern machine learning models, especially large language models (LLMs), employ significantly large model architectures and train on massive datasets. At this scale, it is infeasible to properly curate the training data, and sensitive or undesirable information (e.g., personally identifiable information, copyrighted material or toxic text) may be ingested by the model during training~\cite{carlini2021extracting,lehman2021bert,carlini2023quantifying, carlini2023poisoning}. Moreover, when models are trained on data collected from individual users, e.g., medical data, some users may request their data to be deleted following the right to be forgotten provided by recent privacy legislation~\cite{voigt2017eu,pardau2018california,act2000personal}. 

Motivated by the above scenarios, \textit{machine unlearning} has emerged as a subfield of machine learning. The goal of machine unlearning is to remove the influence of a specific subset of training examples from a trained model, while maintaining the performance of the model. A straightforward machine unlearning method is to retrain the model on an updated training set that excludes the samples to be removed. However, retraining deep models, especially LLMs, from scratch is infeasible due to the exorbitant computational costs. Several unlearning techniques have recently been proposed to tackle this challenge by efficiently removing the influence of the data to be unlearned (see Appendix~\ref{sec:related_work} for an overview). 

Even though machine unlearning has recently received significant research attention, implications of unlearning on other crucial aspects such as fairness have been scantly explored. Fairness is especially critical for language models, since these models are embedded in a variety of applications including call centers and other question-answer applications where the output may jeopardize people's chances to obtain services or may lead to unfair treatment. Several works have demonstrated the bias of language models, e.g., \cite{bolukbasi2016man,borkan2019nuanced,hutchinson2020social,devassimonmanela2021stereotype,baldini2022your}. However, there has been little understanding on the effects of unlearning on the overall fairness of the model.

In this work, we evaluate the fairness of LLMs trained using a popular unlearning framework, called Sharded, Isolated, Sliced, and Aggregated (SISA) Training \cite{bourtoule2021machine}. The SISA framework partitions the training data into disjoint shards, and trains a constituent model on each shard, thus creating an ensemble of models. During inference, predictions from the constituent models are aggregated, typically using majority voting. During unlearning, only impacted constituent model(s) needs to be retrained on much smaller shard(s), resulting in significant speed ups. Our choice of SISA among unlearning techniques is motivated by the following reasons. First, SISA is an \textit{exact} unlearning framework with certifiable guarantees. In contrast, several unlearning methods provide only \textit{approximate} unlearning (see Section~\ref{sec:preliminaries} for details). Second, the SISA framework can be applied to a wide variety of model architectures including transformer-based language models. On the contrary, several unlearning methods, especially with certified guarantees, are not applicable to language models (e.g., \cite{sekhari2021remember,warnecke2023machine}). Overall, SISA can comply with regulations that require data removal at the requests of the data owner.

To improve the fairness of LLMs trained using SISA, we propose to apply \textit{post-processing techniques} for bias mitigation. Unlike \textit{pre-processing} and \textit{in-processing} bias mitigation techniques, post-processing techniques do not require any modification of the training data or model training procedures, and they only modify the model output (see \cite{bellamy2019ai} for a summary). Given the massive costs associated with training large LLMs and the vast amount of unlabeled data used during this process, we argue that post-processing approaches can minimize the environmental impact of re-training LLMs. 

\textbf{Contributions:} To the best of our knowledge, this is the first work to explore the effects of unlearning on the fairness of LLMs. 
We outline our contributions below.
\vspace{-5pt}
\begin{itemize}[leftmargin=*]
\itemsep0em
\item We study the accuracy-fairness trade-off of LLMs trained using the SISA unlearning framework by focusing on the task of toxic text classification. We measure model bias in terms of \textit{group fairness} using the notion of \textit{equalized odds}, by following the setup in \cite{baldini2022your}. We empirically demonstrate that the SISA framework can produce models that are less fair.
\item We investigate post-processing bias mitigation techniques in the context of SISA ensembles to improve the accuracy-fairness trade-off. We adapt the post-processing method from \cite{hardt2016equality} to design three methods that can handle model ensembles. We prove that the third method generalizes the post-processing optimization problem in \cite{hardt2016equality} for ensemble of models and produces an optimal fair predictor, which is our key theoretical contribution and can be of independent interest outside of SISA unlearning. We empirically evaluate the three post-processing methods for SISA on three state-of-the-art LLM architectures. 
\end{itemize}

\section{Preliminaries}\label{sec:preliminaries}

\textbf{Machine Unlearning:}
Supervised machine learning is a process to learn a \textit{model}, in particular, a parameterized function $f_{\mathbf{w}}$ that, given an input from input space $\mathcal{X}$, can predict an output from output space $\mathcal{Y}$, i.e., $f_{\mathbf{w}}: \mathcal{X} \to \mathcal{Y}$. The parameters are typically optimized by applying methods such as stochastic gradient descent (SGD) to a training set. Let $\mathcal{D}$ be a fixed training dataset consisting of $N$ samples, i.e., $\mathcal{D} = \{\mathbf{z}_1,\mathbf{z}_2,\ldots,\mathbf{z}_N\}$, where each $\mathbf{z}_i = (\mathbf{x}_i, \mathbf{y}_i)\in\mathcal{X}\times\mathcal{Y}$. Let us abstract out the training process using a (randomized) algorithm $\mathcal{A}$ that trains on $\mathcal{D}$ and outputs a model $\mathbf{w}\in\mathcal{W}$, where $\mathcal{W}\subseteq\mathbb{R}^d$ denotes the parameter space of a hypothesis class. 
Note that randomness in $\mathcal{A}$ induces a probability distribution over the models in the parameter space.

An unlearning algorithm $\mathcal{U}$ takes as input the trained model $\mathbf{w}= \mathcal{A}(\mathcal{D})$, data to be unlearned $\mathcal{D}^u\subset\mathcal{D}$ and retain dataset $\mathcal{D}^r = \mathcal{D}\setminus\mathcal{D}^u$, and outputs a new model $\mathbf{w}^u$, i.e., $\mathbf{w}^u = \mathcal{U}(\mathcal{A}(\mathcal{D}),\mathcal{D}^u,\mathcal{D}^r)$.
\textit{Exact unlearning} methods essentially ensure that the distribution of the unlearned model $\mathcal{U}(\mathcal{A}(\mathcal{D}),\mathcal{D}^u,\mathcal{D}^r)$ is perfectly indistinguishable from that of the retrained model $\mathcal{A}(\mathcal{D}^r)$ \cite{bourtoule2021machine}. On the other hand, \textit{approximate unlearning} methods ensure that the distributions of unlearned and retrained models are stochastically indistinguishable, where stochastic indistinguishability is typically characterized by using notions similar to differential privacy~\citet{guo2019certified,sekhari2021remember,warnecke2023machine}. 

\textbf{Sharded, Isolated, Sliced, and Aggregated (SISA) Training:}
\cite{bourtoule2021machine} proposed SISA, an exact unlearning method that reduces the computational overhead associated with retraining from scratch. The SISA framework randomly divides the training dataset $\mathcal{D}$ into $S$ disjoint shards $\mathcal{D}_1, \ldots, \mathcal{D}_S$ of approximately equal size. During training, for each shard $\mathcal{D}_k$, a \textit{constituent model}, denoted as $M_k$, is trained. The data in each shard is further partitioned into $R$ disjoint slices, where each constituent model is trained incrementally on each slice and the model parameters after training on a slice are saved. At inference time, $S$ individual predictions from the constituent models are aggregated, typically, through majority voting (similar to ensemble methods~\cite{dietterich2000ensemble}). 

When one or more data samples need to be unlearned, only the constituent models corresponding to the shards that contain the data sample(s) are retrained. More specifically, the slice containing the data sample(s) to be unlearned and the following slices in the same shard need to be retrained. Towards this end, the last saved checkpoint before including the slice containing the data sample(s) to be unlearned can be used as a starting point. This provides significant speed ups over conventional retraining of a single model.

\textbf{Fairness for Toxic Text Classification:}
We consider the task of toxic text classification, and measure model bias in terms of \textit{group fairness}~\cite{chouldechova2018the} by following the setup in~\cite{baldini2022your}. In particular, we consider certain topics, such as religion or race, as sensitive. If a text sample contains one or more terms related to a sensitive topic (e.g., religion), we say that it belongs to a \textit{sensitive group}; otherwise, to the complementary group (no religion). We analyze the fairness of toxic text prediction in the presence or absence of sensitive information (e.g., religion or race), with the goal that the performance of a fair predictor should not be influenced by these sensitive topics. 

While there are several notions of group fairness, e.g., demographic parity (see \cite{verma2018fairness, czarnowska2021quantifying}), we consider the notion of \textit{equalized odds}~\cite{hardt2016equality}. Essentially, equalized odds requires that the model output conditioned on the true label to be independent of the sensitive attribute. More formally, let $Y$ denote the true label (e.g., toxic text), $X$ denote the features, and $A$ denote the sensitive attribute (e.g., religion or race). Let $\hat{Y} = f_{\mathbf{w}}(X,A)$ be the model output, denoted as the \textit{predictor}. Equalized odds requires that the model predictor $\hat{Y}$ has equal \textit{true positive rates} and \textit{false positive rates} across the privileged (i.e., sensitive) and unprivileged (i.e., non-sensitive) groups, satisfying the following constraint:
\vspace{-4pt}
\begin{equation}
    \label{eq:EO}
    \Pr\left(\hat{Y}=1\mid A=0, Y=y\right) = \Pr\left(\hat{Y}=1\mid A=1, Y=y\right), \quad y\in\{0,1\}.
\end{equation}
\vspace{-6pt}

\textbf{Baseline Post-Processing Method for Fairness:}
To improve the model fairness without retraining, we explore the use of post-processing methods. We build on the post-processing method proposed in \cite{hardt2016equality}, who originally proposed the notion of equalized odds. We denote the method as \textit{HPS}, using the last names of the authors. 

The HPS method constructs a \textit{derived predictor} $\tilde{Y}$, which only depends on the predicted label $\hat{Y}$ and the sensitive attribute $A$, and satisfies equalized odds while minimizing classification loss. Specifically, let $\ell : \{0, 1\}^2 \to \mathbb{R}$ denote a loss function that takes a pair of labels and returns a real number. Let us define $p_{ya} = \Pr\left(\tilde{Y} = 1 \mid \hat{Y} = y, A = a\right)$. Then, the HPS method constructs $\tilde{Y}$ by solving the following optimization problem:
\vspace{-2pt}
\begin{IEEEeqnarray}{rCl}
    \min_{p_{ya}} & \quad & \mathbb{E}\left[\ell(\tilde{Y}, Y)\right]\\
    \textrm{s.t.} & \quad & \Pr\left(\tilde{Y} = 1 \mid A = 0, Y = 0 \right) = \Pr\left(\tilde{Y} = 1 \mid A = 1, Y = 0 \right),\nonumber\\
    & \quad & \Pr\left(\tilde{Y} = 1 \mid A = 0, Y = 1 \right) = \Pr\left(\tilde{Y} = 1 \mid A = 1, Y = 1 \right),\nonumber\\
    & \quad & 0 \leq p_{ya} \leq 1.\nonumber
\end{IEEEeqnarray}
One can show that the above optimization problem is a linear program in four variables $\{p_{ya}: y\in\{0,1\}, a\in\{0,1\}\}$~\cite{hardt2016equality}. We denote the derived predictor obtained by solving the above optimization problem as $\textsc{HPS}(\hat{Y})$. Next, we adapt the HPS method to design post-processing methods for the ensemble of models produced by SISA.

\section{FairSISA: Ensemble Post-Processing for SISA}\label{sec:fairsisa}

Let $\hat{Y}_1, \hat{Y}_2, \dots, \hat{Y}_S$ denote the predictions from the SISA constituent models. We consider three ways to perform post-processing for SISA. 

\textbf{Aggregate then post-process:} The most natural way to apply post-processing to SISA is after aggregating the predictions from the constituent models. We focus on the majority voting aggregation rule, since it is demonstrated to perform well \cite{bourtoule2021machine}. We denote majority voting as
\vspace{-2pt}
\begin{equation}
    \label{eq:majority-vote}
    \textsc{Maj}\left(\hat{Y}_1, \hat{Y}_2, \dots, \hat{Y}_S\right) = \argmax_{y\in\{0,1\}}\: n_y, \quad \textrm{where } n_y = \left|\{i\in[S] : \hat{Y}_i = y\}\right|.
\end{equation}
Then, the derived predictor obtained by first aggregating and then post-processing can be defined as $\textsc{HPS}\left(\textsc{Maj}\left(\hat{Y}_1, \hat{Y}_2, \dots, \hat{Y}_S\right)\right)$.

\textbf{Post-process then aggregate:} Another natural way to apply post-processing to SISA is to first post-process the predictions from each constituent model and then aggregate the post-processed predictions. Again, focusing on the majority voting aggregation rule, the derived predictor obtained by first post-processing and then aggregating can be defined as $\textsc{Maj}\left(\textsc{HPS}(\hat{Y}_1), \textsc{HPS}(\hat{Y}_2), \dots, \textsc{HPS}(\hat{Y}_S)\right)$.

\textbf{Ensemble post-processing:} Instead of aggregating the predictions before or after post-processing with a specific aggregation rule (such as majority voting), we design a post-processing method that can inherently aggregate the predictions. In particular, we generalize the HPS optimization problem to handle ensemble predictions. Recall that $\ell : \{0, 1\}^2 \to \mathbb{R}$ denotes a loss function that takes a pair of labels and returns a real number. For a length-$S$ binary vector $\bar{y}\in\{0,1\}^S$ and $a\in\{0,1\}$, let us define 
$p_{\bar{y}a} = \Pr\left(\tilde{Y} = 1 \mid \hat{Y}_1 = \bar{y}_1, \hat{Y}_2 = \bar{y}_2, \dots, \hat{Y}_S = \bar{y}_S, A = a\right)$.
We propose an ensemble post-processing method that constructs $\tilde{Y}$ by solving the following optimization problem:
\begin{IEEEeqnarray}{rCl}
\label{eq:ensemble-pp}
\min_{p_{\bar{y}a}} & \quad & \mathbb{E}\left[\ell(\tilde{Y}, Y)\right]\\
\textrm{s.t.} & \quad &
\Pr\left(\tilde{Y}=1 \mid A = 0, Y = 0\right) = \Pr\left(\tilde{Y}=1 \mid A = 1, Y = 0\right),\nonumber\\
& \quad & 
\Pr\left(\tilde{Y}=1 \mid A = 0, Y = 1\right) = 
\Pr\left(\tilde{Y}=1 \mid A = 1, Y = 1\right),\nonumber\\
& \quad &
0\leq p_{\bar{y}a} \leq 1\nonumber.
\end{IEEEeqnarray}

Next, we show that the above optimization problem is a linear program, which produces an optimal derived predictor for the ensemble of models. The proof is deferred to Appendix~\ref{sec:proof}

\begin{proposition}
    \label{prop:LP}
    The optimization problem in~\eqref{eq:ensemble-pp} is a linear program in $2^{S+1}$ variables $\{p_{\bar{y}a} : \bar{y}\in\{0,1\}^S, A\in\{0,1\}\}$, whose coefficients can be computed from the joint distribution of $(\hat{Y}_1, \hat{Y}_2, \dots, \hat{Y}_S, A, Y)$. Further, its solution is an optimal equalized odds predictor derived from $\hat{Y}_1, \hat{Y}_2, \dots, \hat{Y}_S$ and $A$.
\end{proposition}

\section{Evaluation}\label{sec:evaluation}

We perform empirical evaluation on using three state-of-the-art models (BERT, DistilGPT2, GPT2) on a representative dataset (HateXplain). For the sake of brevity, in the main paper, we discuss empirical results for BERT and DistilGPT2. Additional details and empirical evaluations are in Appendix~\ref{sec:additional-evaluation}. 

HateXplain is a benchmark hate speech dataset which consists of 20K posts from Twitter and Gab~\cite{mathew2021hatexplain}. The dataset has fine-grained annotations for religion, race, and gender. We use coarse-grained groups as sensitive groups (e.g., mention of any religion) as opposed to the finer-grained annotations (e.g., Hindu), similar to \cite{baldini2022your}. This is because, for HateXplain, most subgroups account for significantly less proportion of the data, and there is considerable overlap between subgroups. We focus on two sensitive attributes: religion and race. We combine the annotations for offensive and hate speech into one class of toxic text, similar to \cite{baldini2022your}. 

\begin{figure}[ht!]
    \centering
    \begin{subfigure}[Religion]{\includegraphics[scale=0.425]{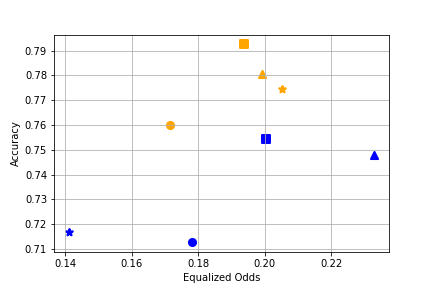}}
    \end{subfigure}
    \begin{subfigure}[Race]{\includegraphics[scale=0.425]{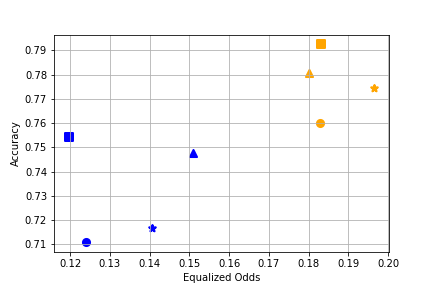}}
    \end{subfigure}\\
    \begin{subfigure}[Legend]{\includegraphics[scale=0.3]{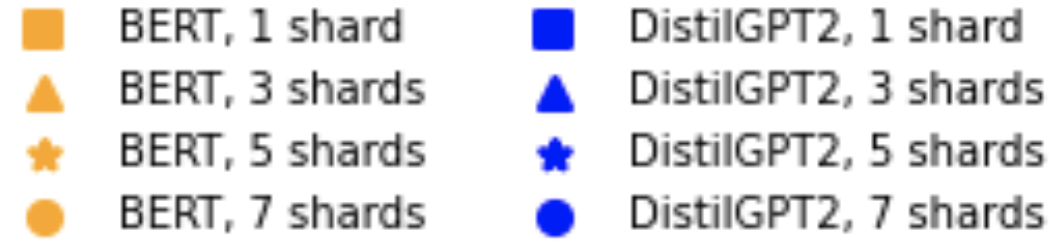}}
    \end{subfigure}
    \caption{Accuracy-fairness trade-off for SISA framework.}
    \label{fig:acc_vs_fairness_SISA}
\end{figure}

\begin{figure}[t!]
    \centering
    \begin{subfigure}{\includegraphics[scale=0.6]{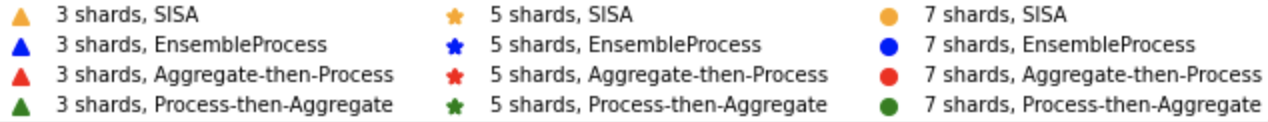}}
    \end{subfigure}\\
    \begin{subfigure}[BERT, Religion]{\includegraphics[scale=0.425]{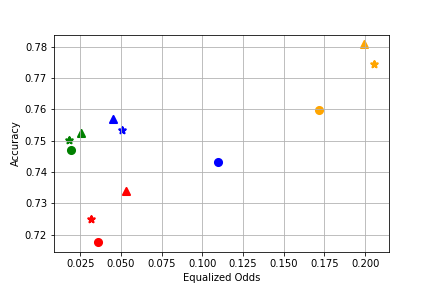}}
    \end{subfigure}
    \begin{subfigure}[BERT, Race]{\includegraphics[scale=0.425]{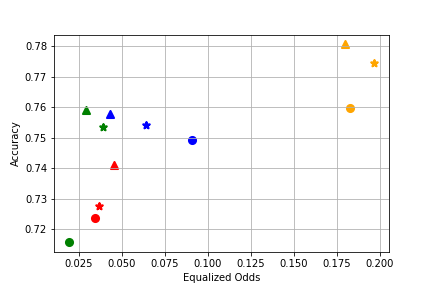}}
    \end{subfigure}\\
    \begin{subfigure}[DistilGPT2, Religion]{\includegraphics[scale=0.425]{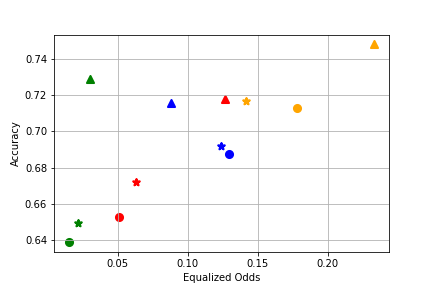}}
    \end{subfigure}
     \begin{subfigure}[DistilGPT2, Race]{\includegraphics[scale=0.425]{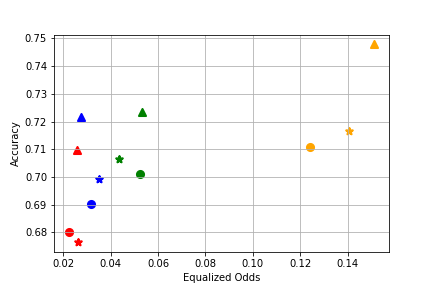}}
    \end{subfigure}
    \caption{Comparison of post-processing methods for SISA.}
    \label{fig:acc_vs_fairness_post_process}
\end{figure}

First, we investigate how SISA training procedure influences the performance-fairness relationship by considering $S= 1, 3, 5,$ and $7$ shards\footnote{We do not consider the \textit{slicing} component of SISA, because, unlike smaller models studied in \cite{bourtoule2021machine}, LLMs incur prohibitively large storage cost for saving model checkpoints for each slice.}. Note that $S=1$ shard corresponds to the conventional single model fine-tuning paradigm. In Figure~\ref{fig:acc_vs_fairness_SISA}, we demonstrate the performance as measured by accuracy on y-axis (higher accuracy is better) and the group fairness as measured by equalized odds (EO) on the x-axis (lower EO is better). We observe that, for both the models and sensitive attributes, the accuracy decreases with the number of shards, which is consistent with the observation in \cite{bourtoule2021machine} for image-domain data. In contrast, EO values vary widely for different number of shards. Importantly, the SISA framework can indeed degrade the fairness (with higher EO values) for both the models and sensitive attributes. For instance, for the BERT model with the Religion attribute (Figure~\ref{fig:acc_vs_fairness_SISA}(a)), SISA with 3 shards consistently result in poorer fairness (higher EO values) than the single model; and for the DistilGPT2 model with the Race attribute (Figure~\ref{fig:acc_vs_fairness_SISA}(c)), SISA results in worse fairness (higher EO values) than the case of conventional single model. These results strongly suggest that it is important to investigate bias mitigation methods for the SISA framework.

Next, we compare the three post-processing methods for bias mitigation from Section~\ref{sec:fairsisa} for the SISA framework. In Figure~\ref{fig:acc_vs_fairness_post_process}, for each model and sensitive attribute, we plot accuracy vs. equalized odds (EO). Amongst the three methods, \textit{Post-process then Aggregate} method generally achieves the best trade-off between the accuracy and EO, whereas \textit{Aggregate then Post-Process} method generally achieves the worst trade-off between the accuracy and EO. The \textit{Ensemble Post-Process} method, in general, achieves the highest accuracy for a moderate EO, which is consistent with the theory that the method is optimal in terms of accuracy (the objective function of the optimization problem \eqref{eq:ensemble-pp}).

\section{Conclusion}\label{sec:conclusion}
We investigated an interplay between unlearning and fairness for LLMs by focusing on a popular unlearning framework called SISA \cite{bourtoule2021machine}. We empirically demonstrated that SISA can indeed reduce fairness in LLMs, measured in terms of a group fairness metric of equalized odds. As a solution, we proposed three post-processing bias mitigation techniques for ensemble models produced by SISA. We theoretically showed that one of the methods generalizes the optimization problem from \cite{hardt2016equality} for ensemble models and produces an optimal derived predictor. We empirically demonstrated the efficacy of our post-processing techniques for SISA.

\section*{Acknowledgements}
This work was supported in part by the European Union's Horizon 2020 research and innovation programme under grant number 951911 – AI4Media.

\bibliography{references}
\bibliographystyle{iclr2024_conference}

\appendix
\section*{Appendix}
\section{Related Work}\label{sec:related_work}

\textbf{Machine Unlearning.} \cite{cao2015towards} were the first to introduce the notion of machine unlearning. Machine unlearning approaches can be divided in two broad categories: (i) exact machine unlearning (e.g., retraining from scratch, SISA~\cite{bourtoule2021machine}) and (ii) approximate machine unlearning \cite{graves2020amnesiac,izzo2021approximate,ginart2019making,golatkar2020eternal,golatkar2020forgetting,thudi2021unrolling}. For details, see recent surveys \cite{nguyen2022survey,xu2023machine}. 

\textbf{AI Fairness.} The goal of AI fairness is to identify and eliminate algorithmic bias from machine learning models. This bias can arise from the difference between individuals or groups with respect to a sensitive attribute (e.g., race, gender, status, etc.). Fairness in machine learning has been widely studied~\cite{bellamy2019ai,zhang2020machine,biswas2020machine,dwork2012fairness,mehrabi2021survey}. Several studies have proposed bias mitigation techniques, which can be grouped in (i) pre-processing~\cite{calmon2017optimized,krasanakis2018adaptive}, (ii) in-processing~\cite{corbett2017algorithmic,zhang2018mitigating}, and (iii) post-processing techniques~\cite{hardt2016equality,pleiss2017fairness,wei2020optimized}. In this work, we focus on post-processing methods because they are directly applied to outputs of the trained model, without requiring to modify the model.

\textbf{Fairness vs Machine Unlearning.} Even though AI fairness has been widely studied and various machine unlearning techniques have been developed, the literature still lacks works studying the impact of machine unlearning techniques. 
\cite{zhang2023forgotten} study fairness properties of different unlearning methods applied to small models. Their results on various tabular datasets show that unlearning can increase disparity. \cite{oesterling2023fair} present the first provably fair unlearning method. They conduct extensive experiments on tabular datasets showing its efficacy. However, this method is restricted to convex functions. \cite{koch2023no} analyze the performance of SISA models in imbalanced datasets. To the best of our knowledge, no work has investigated the impact of SISA models on fairness for LLMs and studied the application of post-processing methods to mitigate it.

\section{Proof of Proposition \ref{prop:LP}}\label{sec:proof}

By the definition of derived predictor $\tilde{Y}$ \cite{hardt2016equality}, it can only depend on ensemble output $\hat{Y}_1, \dots, \hat{Y}_S$ and $A$. Since these variables are binary, the predictor $\tilde{Y}$ is completely defined by the $2^{S+1}$ parameters in $[0,1]$ corresponding to the probabilities $p_{\bar{y}a} = \Pr\left(\tilde{Y} = 1 \mid \hat{Y}_1 = \bar{y}_1, \hat{Y}_2 = \bar{y}_2, \dots, \hat{Y}_S = \bar{y}_S, A = a\right)$. 

Next, we show that the objective function is a linear function in these parameters. For simplicity of notation, we define, $p_{\bar{y}a}^y = \Pr\left(\tilde{Y} = y \mid \hat{Y}_1 = \bar{y}_1, \hat{Y}_2 = \bar{y}_2, \dots, \hat{Y}_S = \bar{y}_S, A = a\right)$. Also, let us denote $\hat{Y}_e$ as the vector of predictions from the $S$ constituent models, i.e., $\hat{Y}_e = [\hat{Y}_1 \: \hat{Y}_2 \: \dots \: \hat{Y}_S]$.

\begin{IEEEeqnarray}{rCl}
\mathbb{E}[\ell(\tilde{Y}, Y)] & = & \sum_{y', y\in\{0,1\}, y'\ne y} \Pr\left(\tilde{Y} = y', Y = y\right)\nonumber\\ 
& = & \sum_{y', y\in\{0,1\}, y'\ne y} \sum_{a\in\{0,1\}, y''\in\{0,1\}^S} \Pr\left(\tilde{Y} = y', Y = y \mid \hat{Y}_e = y'', A = a\right) \Pr\left(\hat{Y}_e = y'', A = a\right)\nonumber\\
& = & \sum_{y', y, y'\ne y} \sum_{a, y''} \Pr\left(\tilde{Y} = y' \mid \hat{Y}_e = y'', A = a\right)\nonumber\\
& & {\quad} \Pr\left(Y = y \mid \hat{Y}_e = y'', A = a\right)
\Pr\left(\hat{Y}_e = y'', A = a\right)\nonumber\\
& = & \sum_{y', y, y'\ne y} \sum_{a, y''} p_{\bar{y}a}^y\Pr\left(Y = y, \hat{Y} = y''\mid A = a\right)
\Pr\left(A = a\right)
\end{IEEEeqnarray}
Since all probability in the last line above that do not involve $\tilde{Y}$ can be computed from the joint distribution, it follows that the objective function is linear in $p_{\bar{y}a}$.

To show that the equalized odds constraints are linear in $p_{\bar{y}a}$, consider the true positive rate (TPR) constraint in the equalized odds definition:

\begin{IEEEeqnarray}{rCl}
\IEEEeqnarraymulticol{3}{l}{\Pr\left(\tilde{Y} = 1 \mid Y = 1, A = a\right)}\nonumber\\ 
& = & \quad \sum_{\bar{y}\in\{0,1\}^S} \Pr\left(\tilde{Y} = 1, \hat{Y}_e = \bar{y} \mid Y = 1, A = a\right)\nonumber\\
& = & \sum_{\bar{y}} \Pr\left(\hat{Y}_e = \bar{y} \mid Y = 1, A = a\right) \Pr\left(\tilde{Y} = 1 \mid Y = 1, \hat{Y}_e = \bar{y}, A = a\right)\nonumber\\
& = & \sum_{\bar{y}} \frac{\Pr\left(\hat{Y}_e = \bar{y}, Y = 1 \mid A = a\right)}{\Pr\left(Y = 1 \mid A = a\right)} \Pr\left(\tilde{Y} = 1 \mid Y = 1, \hat{Y}_e = \bar{y}, A = a\right)\nonumber\\
& = & \sum_{\bar{y}} \frac{\Pr\left(\hat{Y}_e = \bar{y}, Y = 1 \mid A = a\right)}{\Pr\left(Y = 1 \mid A = a\right)} \Pr\left(\tilde{Y} = 1 \mid \hat{Y}_e = \bar{y}, A = a\right)\nonumber\\
& = & \sum_{\bar{y}} \frac{\Pr\left(\hat{Y}_e = \bar{y}, Y = 1 \mid A = a\right)}{\Pr\left(Y = 1 \mid A = a\right)} p_{\bar{y}a}
\end{IEEEeqnarray}
All probabilities in the last line above that do not involve $\tilde{Y}$ can be computed from the joint distribution, and it follows that the TPR is linear in $p_{\bar{y}a}$. Similarly, we can show that the false positive rate portion of the EO constraints is also linear in $p_{\bar{y}a}$. 

Therefore, both the objective function and the constraints of \eqref{eq:ensemble-pp} are linear in $p_{\bar{y}a}$, which completes the proof of the first part of Proposition \ref{prop:LP}. The optimality $\tilde{Y}$ follows from the fact that the optimal solution of \eqref{eq:ensemble-pp} minimizes the loss while satisfying equalized odds, which completes the proof of the second part of Proposition \ref{prop:LP}.

\section{Additional Empirical Evaluations}\label{sec:additional-evaluation}

\subsection{Details on Evaluation Parameters}\label{sec:parameters}
We use the Hugging Face implementation of Transformers \cite{wolf2020transformers} and the corresponding implementations for language models. We use the text sequence classifier without any modifications to increase reproducibility. We fine-tune models for $[S, S+1, S+2]$ epochs, where $S$ is the number of shards ($S=1$ for the baseline single model case), and choose the parameter based on validation accuracy. To tackle any variance, we run experiments for five random seeds and report the average results. 

\subsection{Evaluation of GPT2}
\begin{figure}[ht!]
    \centering
    \begin{subfigure}[Religion]{\includegraphics[scale=0.45]{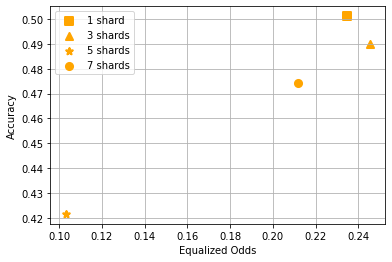}}
    \end{subfigure}
    \begin{subfigure}[Race]{\includegraphics[scale=0.38]{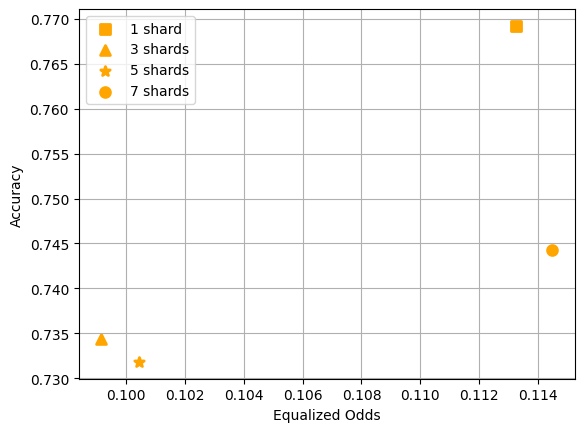}}
    \end{subfigure}
    \caption{Accuracy-fairness trade-off for SISA framework for GPT2.}
    \label{fig:acc_vs_fairness_SISA_GPT2}
\end{figure}

In Figure~\ref{fig:acc_vs_fairness_SISA_GPT2}, we demonstrate the performance as measured by accuracy on y-axis (higher accuracy is better) and the group fairness as measured by equalized odds (EO) on the x-axis (lower EO is better). We observe similar trends as for BERT and DistilGPT2. For both sensitive attributes, the accuracy generally decreases with the number of shards, whereas, EO values vary for different number of shards. For both the attributes, the SISA framework can degrade the fairness (with higher EO values) for GPT2. 

\begin{figure}[ht!]
    \centering
    \begin{subfigure}{\includegraphics[scale=0.5]{figures/camready_legend_PP.png}}
    \end{subfigure}\\
    \begin{subfigure}[GPT2, Religion]{\includegraphics[scale=0.45]{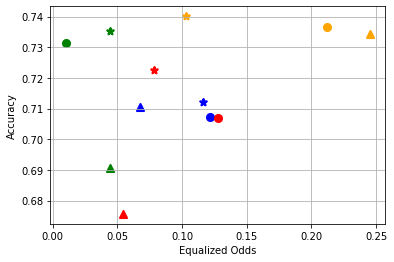}}
    \end{subfigure}
    \begin{subfigure}[GPT2, Race]{\includegraphics[scale=0.45]{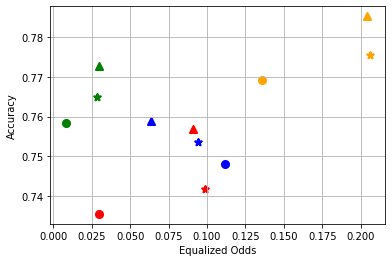}}
    \end{subfigure}
    \caption{Comparison of post-processing methods for SISA for GPT2.}
    \label{fig:acc_vs_fairness_post_process_GPT2}
\end{figure}

Next, we compare three post-processing methods for bias mitigation from Section~\ref{sec:fairsisa} for the SISA framework. In Figure~\ref{fig:acc_vs_fairness_post_process_GPT2}, we plot accuracy vs. equalized odds (EO). Again, we observe similar trends as for BERT and DistilGPT2. Amongst the three methods, Post-process then Aggregate method generally achieves the best trade-off between the accuracy and EO.

\subsection{One Fair Shard}\label{sec:fair_shard}

Next, we study the scenario where one shard is fair and the others are unfair. To obtain a fair shard, we get an equal number of data samples from each possible combination of the values of the sensitive attribute (religion or non-religion) and the labels (toxic or normal). We split the remaining data samples randomly between the other shards. We note that, if the unlearning likelihood of individual data sample is known or can be estimated, SISA can place data samples with high unlearning likelihood on designated shards. This can indeed result in some shards being unfair. 

\begin{figure}
    \centering
    \includegraphics[scale=0.5]{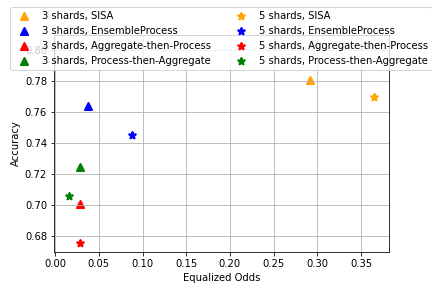}
    \caption{Comparison of post-processing methods for SISA for BERT when one shard is fair and the others are unfair.}
    \label{fig:fair_shard_PP}
\end{figure}

We compare the post-processing methods for $S = 3$ and $5$ shards for the sensitive attribute of religion in Figure~\ref{fig:fair_shard_PP}. First, we observe that the EO values without any post-processing are much larger that the case of random splitting. Post-processing methods significantly reduce model bias, and the \textit{Ensemble Post-processing} methods achieves highest accuracy and substantially low EO, which is consistent with its theoretical optimality.

\end{document}